\documentclass[sigconf]{acmart}
\acmSubmissionID{pos\_141s1}

\usepackage{booktabs} 

\usepackage[ruled]{algorithm2e} 
\usepackage{combelow}
\usepackage{hyperref}  
\usepackage{hyperxmp}  

\SetAlFnt{\small}
\SetAlCapFnt{\small}
\SetAlCapNameFnt{\small}
\SetAlCapHSkip{0pt}

\copyrightyear{2022}
\acmYear{2022}
\setcopyright{rightsretained}
\acmConference{SIGGRAPH ’22 Posters}{August 08-11, 2022}{Vancouver, Canada}
\acmBooktitle{SIGGRAPH ’22 Posters, August 08-11, 2022, Vancouver, Canada}
\acmSubmissionID{pos\_141s1}
\acmDOI{10.1145/3532719.3543208}
\acmISBN{978-1-4503-1234-5/22/07}

\begin{document}
\title{A Fast Text-Driven Approach for Generating Artistic Content}



\author{Marian Lupa\cb{s}cu}
\authornote{Equal contribution, ordered alphabetically.}
\affiliation{%
  \institution{Adobe Research}
  \city{Bucharest}
  \country{Romania}
}
\email{lupascu@adobe.com}

\author{Ryan Murdock}
\authornotemark[1]
\affiliation{%
  \institution{Adobe Research}
  \city{Lehi}
  \country{USA}
}
\email{rmurdock@adobe.com}

\author{Ionu\cb{t} Mironic\u{a}}
\affiliation{%
  \institution{Adobe Research}
  \city{Bucharest}
  \country{Romania}
}
\email{mironica@adobe.com}

\author{Yijun Li}
\affiliation{%
  \institution{Adobe Research}
  \city{Seattle}
  \country{USA}
}
\email{yijli@adobe.com}
\renewcommand\shortauthors{Lupa\cb{s}cu, Murdock, Mironic\u{a}, Li}

%
%

\ccsdesc[500]{Computing methodologies~Learning latent representations}
\ccsdesc[500]{Computing methodologies~Mixture models}

%
%

\keywords{Generative art, image generation, optimization, style transfer, stylization with text, synthetic art}

\begin{teaserfigure}
  \includegraphics[width=\textwidth]{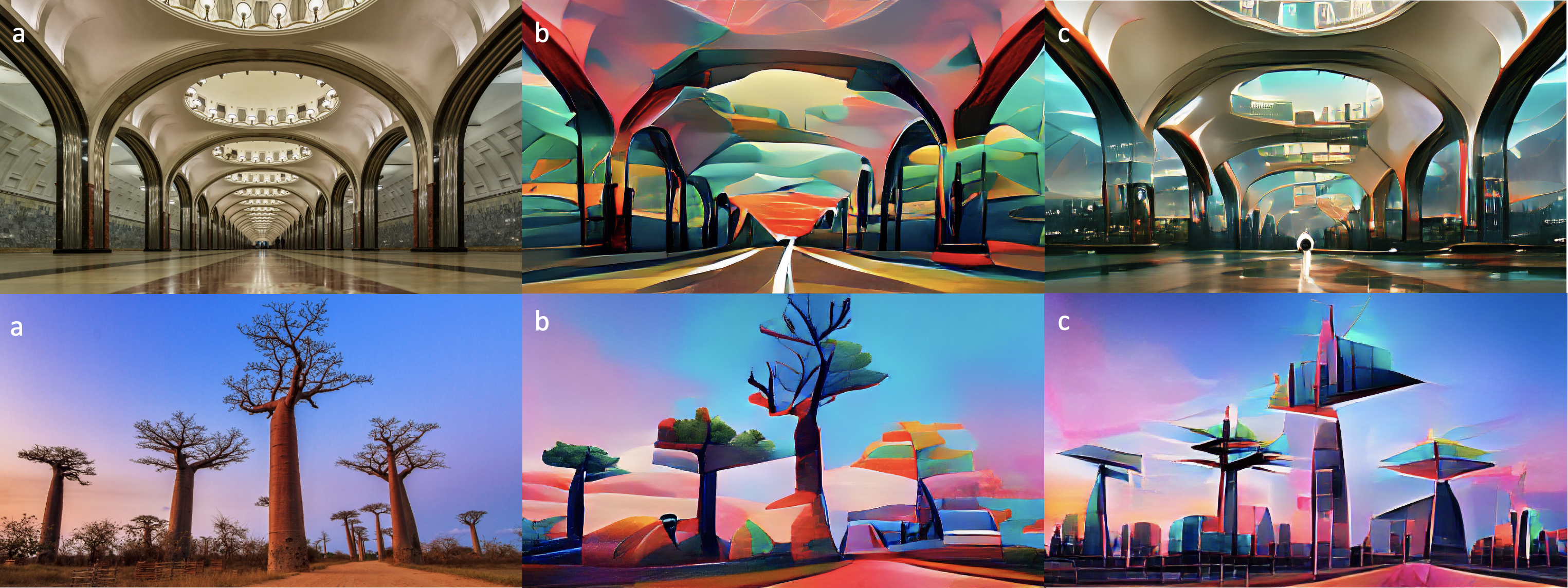}
  \vspace{-20pt}
  \caption{Results of our proposed method, where (a) represents the content image, (b) is the content image stylized with the text "A bold color landscape depicting the road everyone lives on in the style of romanticism paintings | Low poly" and (c) represents the content image stylized with the text "City of the future | Geometric art".}
  \label{fig:teaser}
\end{teaserfigure}

\maketitle

\section{Introduction}
Finding a source of inspiration for visual art creation can be a difficult task. Even with the online search engine, it is still labor-intensive because manually crawling the web is time-consuming.  Therefore, a large number of artistic prototypes must be explored before an artist has a better feel of what their future creations look like structurally but also artistically. In this work, the image can be stylized according to the user’s requirements with a text prompt, a style image, or a combination of style parameters (Figure~\ref{fig:teaser}). The only limitations are the user’s imagination.

\begin{figure*}[t]
  \centering
  \vspace{-5pt}
  \includegraphics[width=1\linewidth]{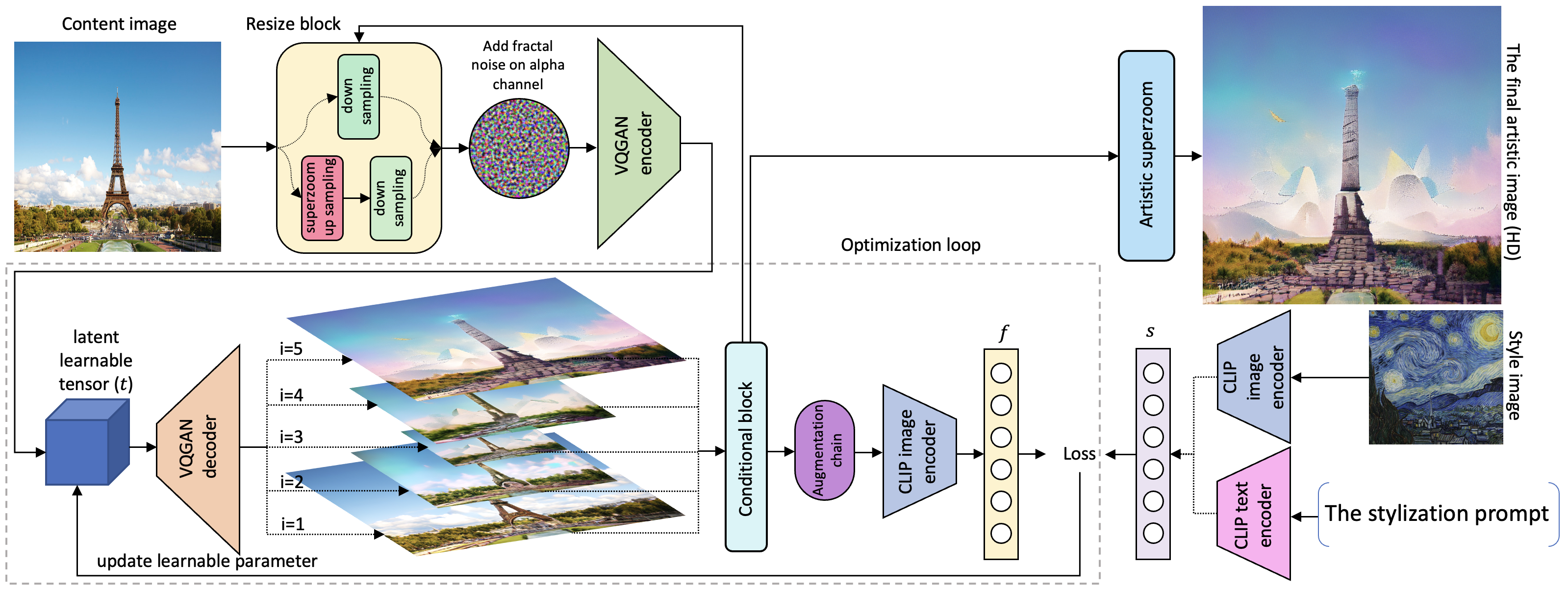}
  \vspace{-15pt}
  \caption{Overview of the proposed stylization pipeline with CLIP and a generative network (VQGAN). The optimization loop is framed in the dotted rectangle. The vector \emph{f} represents the projection in the CLIP space of the image from the current iteration and the vector \emph{s} represents the projection in the CLIP space of a style parameter (image or text).}
  \label{fig:framework}
  \vspace{-5pt}
\end{figure*}

Despite the considerable progress of generative models, there is not a common solution that allows stylization with image and text. The Conditional-driven GAN \cite{pmlr-v48-reed16} first extracted visual details from the text and then synthesized the output image conditioned on these previously extracted features. The Attention-driven GAN \cite{Xu_2018_CVPR} introduced a similarity module between text features and image features. The DALLE \cite{pmlr-v139-ramesh21a} and StyleCLIP \cite{Patashnik_2021_ICCV} are the state-of-the-art approaches of generating and manipulating images via text. However, the DALLE does not support the stylization with a style example and StyleCLIP is mainly limited in generating images from a single domain: faces, cars, etc.

\section{APPROACH}
Our generative network is guided by the CLIP \cite{pmlr-v139-radford21a} to generate images that closely match a description using gradient descent through backpropagation. We use the VQGAN \cite{Esser_2021_CVPR} as the generative network, but it can be replaced with other generative models such as BigGAN \cite{Brock2019LargeSG} and diffusion models \cite{dhariwal2021diffusion}.

The whole pipeline of our proposed method is shown in Figure~\ref{fig:framework}. The first step in the stylization process is represented by the projection of the stylization images and text in the CLIP latent space. 
Also at this step, the learnable tensor from the latent space of the generative model is defined. In our case where the generative model is VQGAN, this tensor $t$ is the encoding of the content image, $t=Encoder(\mathcal{O})$ where $Encoder(\cdot)$ is the VQGAN encoder and $\mathcal{O}$ is the content image (for the case without content images, $t$ is randomly sampled in VQGAN latent space). This step is performed only once regardless of the number of iterations.

A stylization step represents a change to this learnable tensor, which is then decoded by the generative model into an RGB image. This modification is performed as follows: in the first phase, the learnable tensor is passed through the generative model, obtaining an RGB image. In the second phase, this image is projected in the CLIP latent space. The third phase involves calculating the loss function between the projection of the learnable tensor and the projection of the style parameters in CLIP latent space. In the
fourth phase, the gradient with respect to the learnable tensor $t$ can be computed using standard back-propagation. At the end, the learnable parameter is updated by decreasing the partial derivatives (taken from the gradient) of the learnable parameter on each component.  The loss function is: 

\begin{equation}
\label{eqn:01}
\setlength{\abovedisplayskip}{0pt}
\setlength{\belowdisplayskip}{4pt}
\setlength{\abovedisplayshortskip}{0pt}
\setlength{\belowdisplayshortskip}{0pt} 
\mathcal{L}=\frac{2}{|\mathcal{S}|}\sum_{s\in{\mathcal{S}}}\arcsin{(\frac{1}{2}{\|\widehat{CLIP(Decoder(t))}-\widehat{CLIP(s)}\|}_2)}^2,
\end{equation}
where $\widehat{\cdot}$ represents the operation of normalizing a vector, $\mathcal{S}$ represents the set of stylization parameters (images and texts), $CLIP(\cdot)$ is the function that encodes an image or a text in the CLIP latent space, $t$ is the learnable tensor and $Decoder(\cdot)$ is the function that project the learnable tensor in the space of RGB images.

To improve the speed and quality of the results, we use hierarchical scaling to different resolutions, fractal noise combined with an augmentation chain and artistic super-resolution. Hierarchical scaling to different resolutions allows fast styling of intermediate images at lower resolutions than the initial one. Noise fractal added to the intermediate results in combination with an augmentation chain (random resize for each dimension independently, random crop, random perspective, random horizontal flip and random Gaussian noise) removes regular surfaces from the content image, which in turn eliminates the problem of vanishing gradients. Artistic super-resolution increase the resolution of the generated images and create specific paintings effects (e.g., brush strokes).

The average run-time of generating an artistic output of size $2048\times 2048$ starting from a content image or from a random point in the VQGAN latent space is about 4.1s (depending on the stylization parameters) on a single GeForce RTX-3090 GPU.

\section{Conclusion}
In this work, we propose a complete framework that generates visual art. Unlike previous stylization methods that are not flexible with style parameters (i.e., they allow stylization with only one style image, a single stylization text or stylization of a content image from a certain domain), our method has no such restriction. In addition, we implement an improved version that can generate a wide range of results with varying degrees of detail, style and structure, with a boost in generation speed. To further enhance the results, we insert an artistic super-resolution module in the generative pipeline. This module will bring additional details such as patterns specific to painters, slight brush marks, and so on.

%
%
%
%

\bibliographystyle{ACM-Reference-Format}
\bibliography{sample-bibliography}

\end{document}